\documentclass[runningheads]{llncs}

\usepackage{eccv}

\usepackage{eccvabbrv}

\usepackage{graphicx}
\usepackage{booktabs}

\usepackage[accsupp]{axessibility}  

\usepackage{hyperref}

\usepackage{orcidlink}

\usepackage[cjk]{kotex}
\usepackage{colortbl}
\usepackage{comment}
\usepackage{fontawesome5}
\usepackage{arydshln}
\usepackage{wrapfig}
\usepackage{multirow}
\newcommand{\Eq}[1]  {Eq.\ (\ref{eq:#1})}

\newcommand{\Fig}[1] {Figure \ref{fig:#1}}
\newcommand{\Figs}[1]{Figures \ref{fig:#1}}
\newcommand{\Tbl}[1]  {Table \ref{tbl:#1}}
\newcommand{\Tbls}[1] {Tables \ref{tbl:#1}}
\newcommand{\Sec}[1] {Section \ref{sec:#1}}

\definecolor{silver}{rgb}{0.75, 0.75, 0.75}
\definecolor{golden}{rgb}{1.0, 0.87, 0.0}
\definecolor{copper}{rgb}{0.72, 0.45, 0.2}
\definecolor{bronze}{rgb}{0.8, 0.5, 0.2}

\newcommand{\Ours}{RayFusion}

\makeatletter
\def\RemoveSpaces#1{\zap@space#1 \@empty}
\makeatother

\begin{document}

\title{Deep Cost Ray Fusion for\\ Sparse Depth Video Completion} 


\titlerunning{RayFusion}

\author{Jungeon Kim\inst{1}\orcidlink{0000-0003-4212-1970} \and
Soongjin Kim\inst{1}\orcidlink{0000-0001-8142-7062} \and
Jaesik Park\inst{2}\orcidlink{0000-0001-5541-409X} \and 
Seungyong Lee\inst{1}\orcidlink{0000-0002-8159-4271}}

\authorrunning{J. Kim et al.}

\institute{POSTECH, South Korea \and
Seoul National University, South Korea\\
\email{\{jungeonkim,kimsj0302,leesy\}@postech.ac.kr}\\
\email{jaesik.park@snu.ac.kr}}

\maketitle

\begin{abstract}

In this paper, we present a learning-based framework for sparse depth video completion.
Given a sparse depth map and a color image at a certain viewpoint, our approach makes a cost volume that is constructed on depth hypothesis planes. To effectively fuse sequential cost volumes of the multiple viewpoints for improved depth completion, we introduce a learning-based cost volume fusion framework, namely {\em RayFusion}, that effectively leverages the attention mechanism for each pair of overlapped rays in adjacent cost volumes. 
As a result of leveraging feature statistics accumulated over time, our proposed framework consistently outperforms or rivals state-of-the-art approaches on diverse indoor and outdoor datasets, including the KITTI Depth Completion benchmark, VOID Depth Completion benchmark, and ScanNetV2 dataset, using much fewer network parameters.

\end{abstract}

\section{Introduction}

With the benefit of capturing the actual distance, range-sensing devices such as Microsoft Kinect, LiDARs, and Intel RealSense have become increasingly popular.
Notably, recent releases of high-end mobile devices like the iPhone are equipped with LiDARs, reflecting this trend. 
However, these depth sensors often suffer from missing or insufficient depth measurements.
To address the challenge, a variety of learning-based depth completion methods have been proposed~\cite{liao2017parse,fu2020depth,zhong2019deep,wong2021unsupervised,tang2020learning,liu2023mff}. Most of the state-of-the-art depth completion methods use merely a single-view RGB-D image to fill in missing depth values, predominantly focusing on extracting informative multimodal features from an input RGB-D image. 
 
 \begin{figure}[t]
    \centering    
    \includegraphics[width=0.95\linewidth,page=1]{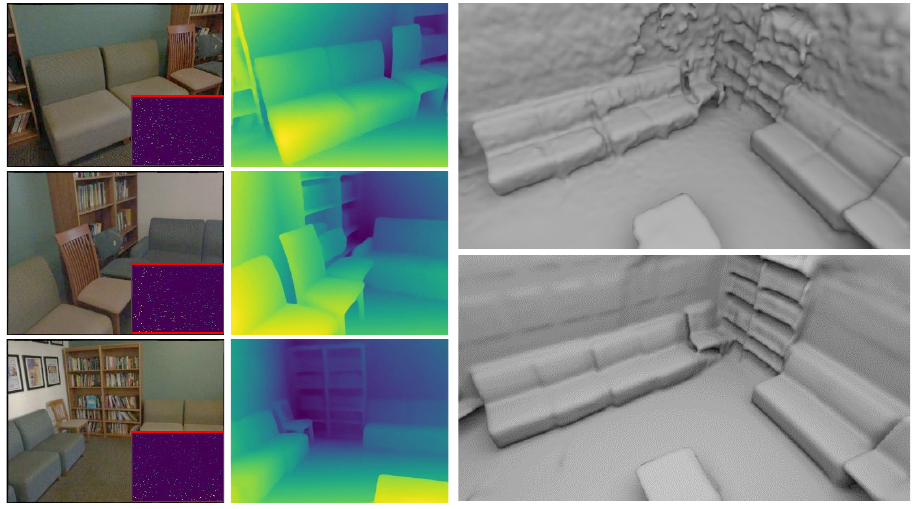}    
    \caption{Depth video completion result of our \Ours{} framework. The framework takes RGB and sparse depth (0.1\% density) video pairs as input (left) and infers completed depth maps (middle). Additionally, we show 3D reconstructions using raw sparse depths (top right) and the completed depths (bottom right). See the supplementary video for various video depth completion results.
    }    
    \label{fig:teaser}
    
\end{figure}

Given the accessibility of RGB-D video data, self-supervised depth completion studies utilized multiple RGB frames for auxiliary photometric loss~\cite{wong2021unsupervised,wong2020unsupervised} for the network training. For more direct utilization of temporal information for enhanced depth completion at test time, a few recent studies~\cite{patil2020don,Krishna_2023_CVPR} tried to fuse feature maps of individual frames using a convolutional long short-term memory (ConvLSTM)~\cite{shi2015convolutional} or spatio-temporal convolution~\cite{tran2018closer}. 
These feature fusion methods basically need warping the previous feature map to align it with the current feature map. However, the alignment is error-prone because such a warping depends on the predicted depths of the previous frame. Although achieving better temporal smoothness, these methods~\cite{patil2020don,Krishna_2023_CVPR} tend to exhibit inferior accuracy compared to the single-view completion methods.

\begin{figure}[t]
    \centering    
    \includegraphics[width=0.99\linewidth,page=1]{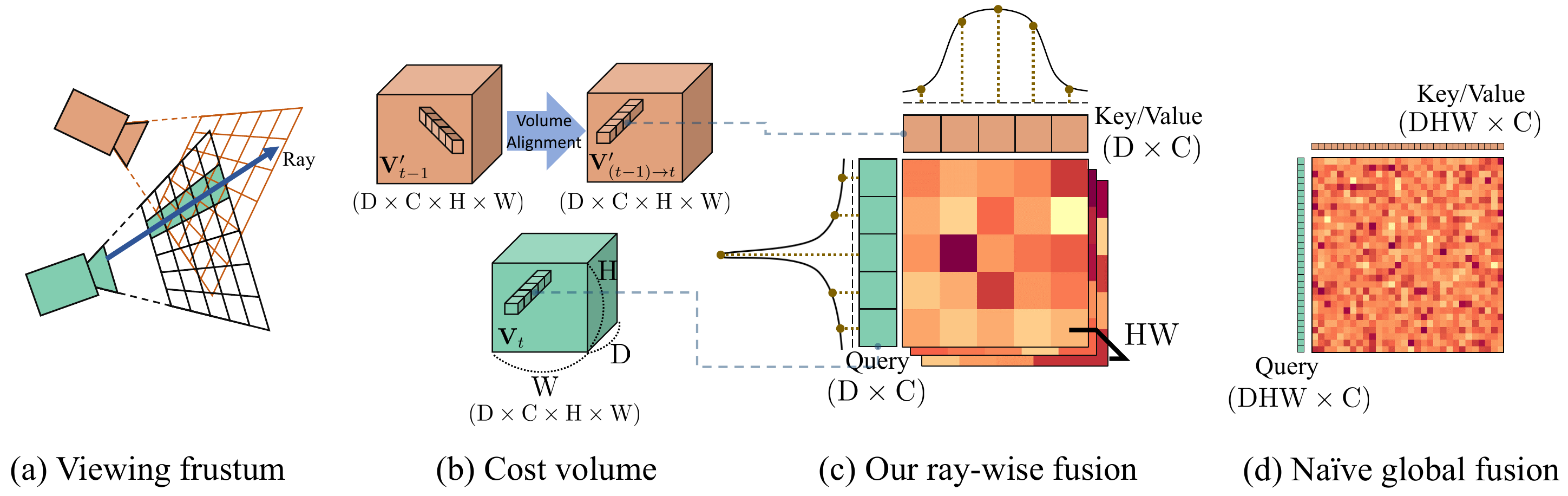}    
    \caption{Illustration of the proposed cost volume fusion scheme. A cost volume is constructed on $D$ depth hypothesis planes and each voxel contains a $C$-dimensional feature vector. When fusing two aligned cost volumes ($\mathbf{V}'_{(t-1)\rightarrow t},\mathbf{V}_{t}$) (b), the proposed scheme (c) applies the attention mechanism into feature sequences corresponding to rays. It is computationally- and memory-efficient than the naive approach (d) of calculating the attention for all features in cost volumes.
    }    
    \label{fig:concept}    
\end{figure}

In this paper, instead of the feature map alignment approach, we utilize a cost volume~\cite{kam2022costdcnet,yao2018mvsnet,yang2020cost,im2019dpsnet}, which has been widely adopted for multi-view stereo, for temporal fusion (\Fig{teaser}). A cost volume is computed with hypothesis depth planes and contains information about probability distributions used for subsequent depth regression. It spatially spans the viewing frustum in the Euclidean space (\Fig{concept} (a)), enabling fusion for cost volumes to be directly performed in the 3D overlapped region of viewing frustums through volume resampling. Therefore, unlike feature image fusion methods, the approach remains unaffected by erroneous depth predictions.

To effectively fuse cost volumes obtained from an RGB-D video, a potential approach is to apply a recurrent neural network (RNN)~\cite{hochreiter1997long,cho2014learning} to overlapped voxels in the cost volume. 
However, this fusion approach can overlook global attributes contained in the cost volumes (\Tbl{kitti_ablation} (\romannumeral5)). The attention mechanism~\cite{vaswani2017attention} could be a good alternative, but applying a global attention scheme to the entire cost volume requires a huge memory footprint and computation resource (\Fig{concept} (d)).

This paper introduces a framework that utilizes a \emph{ray-based cost volume fusion
scheme}. Consider a ray that penetrates two cost volumes of different viewpoints (\Fig{concept} (a)). Our fusion scheme is basically motivated by observation that the features along the ray within cost volumes contain information about probability distributions on hypothesis depth planes. We make the ray-wise features (\Fig{concept} (b)) from two views become a minimal unit for the volume fusion, avoiding a heavy memory footprint, unlike whole volume attention.
Our fusion procedure for ray-wise features comprises two sequential stages: self-attention for refining a current-view depth hypothesis and cross-attention for fusing current-view and previous-view cost volumes. We employ the cross entropy (CE) loss to effectively train the fusion module using pseudo ground truth probability distributions~\cite{Nuanes_2021_CVPR}.

We validate the proposed framework through comprehensive experiments on diverse indoor and outdoor datasets, including the KITTI Depth Completion benchmark~\cite{Uhrig2017THREEDV}, VOID Depth Completion benchmark~\cite{wong2020unsupervised}, and ScanNetV2 dataset~\cite{dai2017scannet}. As a result, we demonstrate outperforming performance over state-of-the-art (SOTA) depth completion methods in both depth and 3D reconstruction metrics and generalization ability despite utilizing significantly fewer network parameters (1.15M parameters - 94.5\% smaller than LRRU~\cite{LRRU_ICCV_2023}) thanks to our effective ray-wise attention design. More interestingly, we demonstrate that the proposed framework, despite not utilizing multiview information (i.e., only using the self-attention stage), still achieves SOTA performance on VOID and ScanNetV2 datasets. This achievement is attributed to our self-attention stage, which refines the cost volume by considering intrinsic properties such as the entropy of probability distributions within the cost volumes.

To summarize, our contributions are as follows:
\begin{itemize}
\item We propose an end-to-end deep learning-based framework, \textit{\Ours{}}, that effectively utilizes temporal information from an input RGB-D video to enhance sparse depth completion. 

\item We propose a novel {\em ray-based cost volume fusion} scheme that leverages the attention mechanism of the Transformer~\cite{vaswani2017attention}
to consider attributes of probability distributions within cost volumes.

\item Our \textit{\Ours{}} consistently outperforms or competes with previous SOTA depth completion methods on various indoor and outdoor datasets with significantly fewer network parameters.

\end{itemize}

\section{Related Work}

Our framework is closely related to multi-view stereo and depth completion research. We review representative methods of those fields.

\noindent\textbf{Multi-view stereo.}
With the advent of deep learning, many multi-view stereo (MVS) methods using deep neural networks have been proposed to replace traditional MVS approaches~\cite{schoenberger2016mvs}. Inspired by the plane sweep stereo, the mainstream deep learning-based methodology in MVS is basically composed of three main stages~\cite{yao2018mvsnet}; image feature extraction, cost volume creation, and cost regularization. Numerous studies have tried to improve those stages or craft novel loss functions to predict accurate depths~\cite{im2019dpsnet,yang2020cost, yao2019recurrent,yang2020cost,gu2020cascade,cheng2020deep, wang2021patchmatchnet,sormann2023dels}. Recently, a few studies~\cite{xi2022raymvsnet,wang2022mvster,ding2022transmvsnet} have employed the attention approach of Transformer for global feature matching on epipolar lines in the image space.

As another line of research, methods that use a monocular RGB video as the input~\cite{hou2019multi,liu2019neural,deepvideomvs_2021_CVPR} have been proposed. Unlike the common MVS studies that use multi-view RGB images with proper baselines as the input, they fully leverage the RGB sequence by the temporal fusion of various representations, including feature images~\cite{deepvideomvs_2021_CVPR}, a latent vector without spatial information~\cite{hou2019multi}, and a probability volume~\cite{liu2019neural} using different techniques (ConvLSTM~\cite{deepvideomvs_2021_CVPR}; the nonparametric Gaussian process~\cite{hou2019multi}; Bayesian filtering implemented as a naive 3D CNN~\cite{liu2019neural}).

\noindent\textbf{Depth completion.} 
Early studies in this field achieve depth completion by considering a depth image as an additional RGB image channel and concatenating it along the channel dimension. The resulting image is then fed into 2D convolutional networks~\cite{ma2018sparse,liao2017parse}. Follow-up studies deal with depth images using separate networks for late fusion with RGB features~\cite{fu2020depth,zhong2019deep,tang2020learning,wong2021unsupervised,liu2023mff}. A few studies utilize networks that estimate a surface normal~\cite{zhang2018deep,huang2019indoor,qiu2019deeplidar},  uncertainty~\cite{eldesokey2020uncertainty,teixeira2020aerial,taguchi2023uncertainty,hu2020PENet}, or edge ~\cite{ramesh2023siunet,tao2021dilated} as a local property related to depth information. 

Recent methods consider depth images in 3D space to properly use 3D positional information~\cite{jeon2021abcd,chen2019learning,chen2022depth,huynh2021boosting}. They back-projected a depth map to obtain a point cloud and extract features on the point cloud. The point cloud features are projected onto the image space and concatenated with RGB features. A few methods adopt Vision Transformer as the backbone to fully leverage the global context on the image domain at the expense of huge network parameters~\cite{Rho_2022_CVPR,zhang2023completionformer}. Unlike aforementioned approaches that focus on extracting better multimodal (2D and 3D) image features, CostDCNet~\cite{kam2022costdcnet} forms a multi-modal feature volume in 3D space, called a cost volume, from a single RGB-D image and infers a completed depth from the cost volume.

The depth completion methods with state-of-the-art (SOTA) performances mainly focus on using only a single-view RGB-D image. While a few studies~\cite{patil2020don,Krishna_2023_CVPR} tried to utilize multiple RGB-D frames by temporally fusing image features via ConvLSTM or spatio-temporal convolution, their results may exhibit inferior accuracy than single-view approaches since their fusion scheme does not adequately address misalignments of adjacent images.

In this paper, we propose an effective framework that fuses two temporally adjacent cost volumes of different viewpoints to infer a more accurate completed depth map. Unlike existing RGB-D video-based methods, our framework performs cost volume fusion in 3D space, which does not rely on the previous erroneous depth prediction. For efficient and effective fusion, we propose a ray-based fusion scheme that leverages the attention mechanism of Transformer~\cite{vaswani2017attention}.

\begin{figure}[t]
	\centering
	\includegraphics[width=0.99\textwidth]{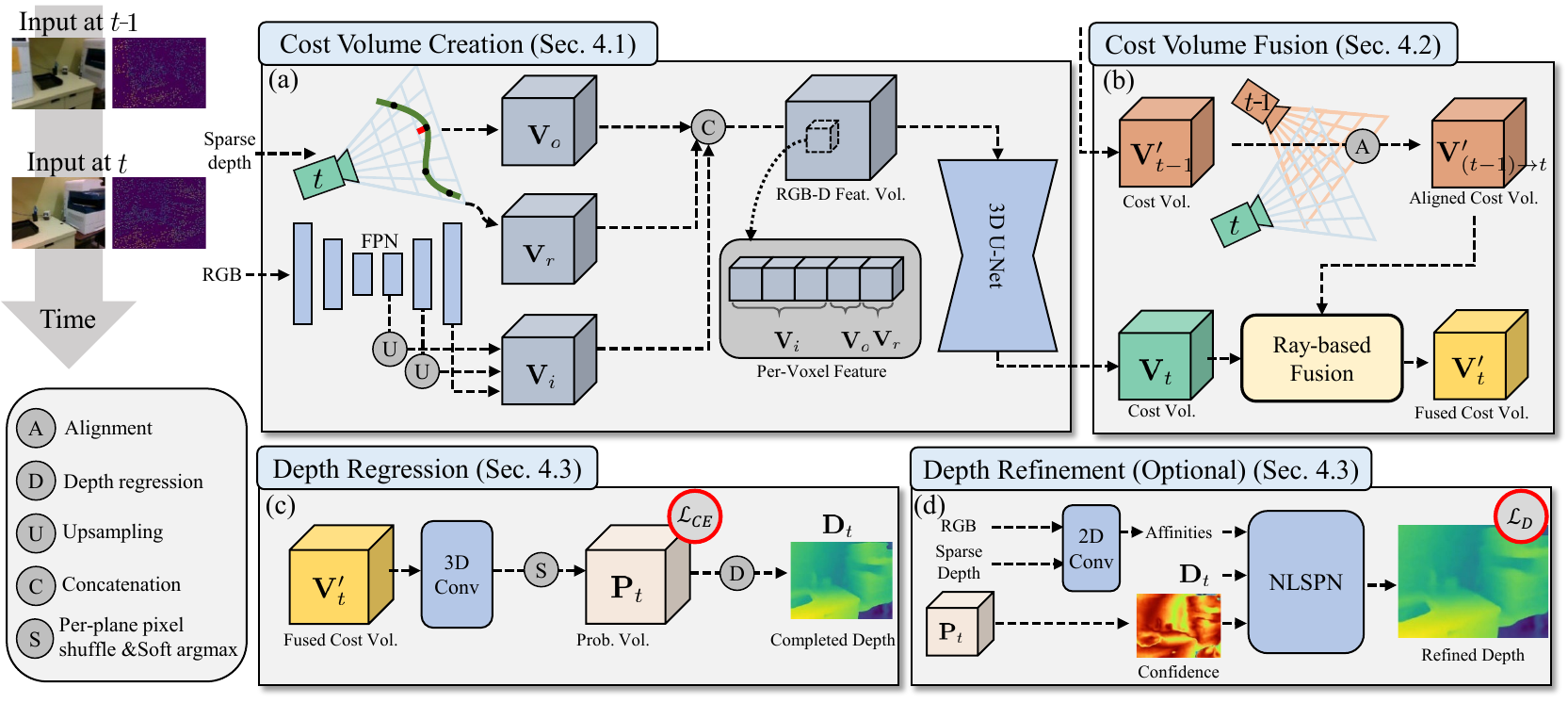}\\        
	\caption{Overall pipeline of our framework. For each frame, our framework infers a cost volume from a single-view RGB-D image (\Sec{cost_vol_gen}) and then fuses the cost volume with the cost volume updated up to the previous frame (\Sec{cost_fusion}). The fused cost volume is used for completed depth regression (\Sec{regress}) and becomes the cost volume for fusion at the next frame. Finally, the completed depth is refined by non-local spatial propagation networks (NLSPN).
	}    
	\label{fig:overview}
\end{figure}

\section{Overview}
For depth video completion with a calibrated camera, we formulate the supervised learning problem as follows: 
\begin{gather} \label{eq:prob_def}
\boldsymbol{\theta}^{\star} = \arg\min_{\boldsymbol{\theta}} \mathcal{L}(\mathbf{D}_t,\mathbf{D}_{gt}), \\
     \mathbf{D}_t = f_{\boldsymbol{\theta}}((\mathbf{I}_{t}, \mathbf{S}_{t}), ..., (\mathbf{I}_{1}, \mathbf{S}_{1}), \mathbf{T}_{t},...,\mathbf{T}_{1}, \mathbf{K}), \nonumber
\end{gather}
where $f_{\boldsymbol{\theta}}$ is a predictor with learnable parameters $\boldsymbol{\theta}$ that uses color $\mathbf{I}_t$ and sparse depth $\mathbf{S}_t$ images, camera poses $\mathbf{T}_t \in SE(3)$, and camera intrinsic parameters $\mathbf{K}$ to infer a completed depth image $\mathbf{D}_t$ at the current frame $t$, $\mathbf{D}_{gt}$ is the ground truth completed depth, and $\mathcal{L}(\cdot,\cdot)$ is a loss function.
To implement the predictor $f_{\boldsymbol{\theta}}$, we propose an incremental cost volume update approach with ray-wise attention, called \textit{RayFusion}, and reformulate the problem as follows:

\begin{gather} \label{eq:prob_def}        
\boldsymbol{\theta}^{\star} = \arg\min_{\boldsymbol{\theta}} \mathcal{L}(\mathbf{D}'_t,\mathbf{D}_{gt}, \mathbf{P}_t, \mathbf{P}_{gt} ), \\
\mathbf{D}'_t= H_{\boldsymbol{\theta}}( \mathbf{D}_t, \mathbf{P}_t, \mathbf{I}_{t}, \mathbf{S}_{t}), \quad \mathbf{D}_t, \mathbf{P}_t = R_{\boldsymbol{\theta}}(\mathbf{V}'_t), \nonumber \\
\mathbf{V}'_t = F_{\boldsymbol{\theta}}(\mathbf{V}'_{t-1}, \mathbf{V}_{t},\mathbf{T}_{t}, \mathbf{T}_{t-1}, \mathbf{K}), \quad
     \mathbf{V}_t = C_{\boldsymbol{\theta}}(\mathbf{I}_{t}, \mathbf{S}_{t}, \mathbf{K}), \nonumber     
\end{gather}
where $C_{\boldsymbol{\theta}}$, $F_{\boldsymbol{\theta}}$, $R_{\boldsymbol{\theta}}$, and $H_{\boldsymbol{\theta}}$ are neural networks for cost volume creation, cost volume fusion, depth regression, and depth refinement, respectively.

For each frame, $C_{\boldsymbol{\theta}}$ predicts a cost volume $\mathbf{V}_t$ using the current RGB $\mathbf{I}_{t}$ and sparse depth $\mathbf{S}_{t}$ images, and camera intrinsics $\mathbf{K}$ as the input.
Then, the predicted cost volume $\mathbf{V}_t$ is fused with the cost volume $\mathbf{V}'_{t-1}$ updated up to the previous frame by $F_{\boldsymbol{\theta}}$. The fused cost volume $\mathbf{V}'_{t}$ at the current frame is used for regressing a completed depth image $\mathbf{D}_t$ via a probability volume $\mathbf{P}_t$ computed by $R_{\boldsymbol{\theta}}$. Lastly, the depth refinement module $H_{\boldsymbol{\theta}}$ improves the completed depth on the image domain using non-local spatial propagation (NLSPN)~\cite{park2020non} to obtain the final completed depth $\mathbf{D}'_t$.

\Fig{overview} shows the overall pipeline of our framework. In the following sections, we elaborate on the main components of our framework, RGB-D cost volume creation (\Sec{cost_vol_gen}), ray-based cost volume fusion (\Sec{cost_fusion}), and completed depth regression and refinement (\Sec{regress}).

\section{Deep Cost Ray Fusion}
\subsection{Cost Volume Creation} \label{sec:cost_vol_gen}
Given an input RGB image and sparse depth samples at each frame, our cost volume creation module $C_{\boldsymbol{\theta}}$ forms occupancy $\mathbf{V}_o$, residual volumes $\mathbf{V}_r$ from sparse depth samples, and RGB feature volume $\mathbf{V}_i$ from multi-scale image features (\Fig{overview} (a)). 
We concatenate these input feature volumes along channel dimensions to obtain an RGB-D feature volume. Then, we feed the RGB-D feature volume into a 3D convolutional U-Net to infer a cost volume at the current frame.
We utilize a modified version of CostDCNet~\cite{kam2022costdcnet} for cost volume creation that does not use a separate geometric feature extractor and uses multi-scale image features.

The feature volume ($\mathbf{V}_o$,$\mathbf{V}_r$,$\mathbf{V}_i$) is constructed on uniformly spaced hypothesis depth planes~\cite{yao2018mvsnet,im2019dpsnet}. The number of depth planes $\textrm{D}$ and the minimum $d_{\textrm{min}}$ and maximum $d_{\textrm{max}}$ depth values of these planes are hyperparameters. When the image spatial resolution is $\textrm{H} \times \textrm{W}$ and the feature dimension is $\textrm{C}$, we have a feature volume $\mathbf{V} \in \mathbb{R}^{D \times C \times H \times W}$. Note that the spatial coverage of the feature volume in 3D Euclidean space corresponds to the viewing frustum at the current frame. More details are described in the supplementary document.

\subsection{Ray-based Fusion}
\label{sec:cost_fusion}
In this section, we explain our ray-based cost volume fusion scheme that effectively considers intrinsic attributes of probability distributions within cost volumes in a memory-efficient manner.
Let us assume that we have a cost volume $\mathbf{V}_{t}$ from the current frame (\Sec{cost_vol_gen}) and a cost volume $\mathbf{V}'_{t-1}$ updated until the previous frame $t-1$, as shown in \Fig{overview} (b). 

\noindent\textbf{Aligning cost volumes.}
We first align two cost volumes ($\mathbf{V}'_{t-1}$, $\mathbf{V}_{t}$) of different viewpoints for the fusion. We utilize the relative camera pose between $t$ and $t-1$ and employ inverse mapping to obtain an aligned cost volume $\mathbf{{V}}'_{(t-1)\rightarrow t}$.
This inverse mapping makes the coordinates of $\mathbf{{V}}'_{t-1}$ to be aligned with the coordinates at the current viewpoint, and it allows easy ray-wise computation in the aligned coordinates. More details can be found in the supplementary document.

\noindent\textbf{Fusion.} We now introduce our approach to fuse two volumes 
$\mathbf{V}_{t}$ and $\mathbf{{V}}'_{(t-1)\rightarrow t}$
using attention mechanism~\cite{vaswani2017attention}. The na\"ive approach is to linearize all voxel features of each cost volume and then compute the cross-attention. However, it requires $\textrm{D}^2\textrm{H}^2\textrm{W}^2$ entries for attention weight calculation, which is impractical due to the huge memory footprint and computation complexity (\Fig{concept} (d)). 

\begin{wrapfigure}{r}{0.40\textwidth}    
    \centering
    \includegraphics[width=0.4\textwidth]{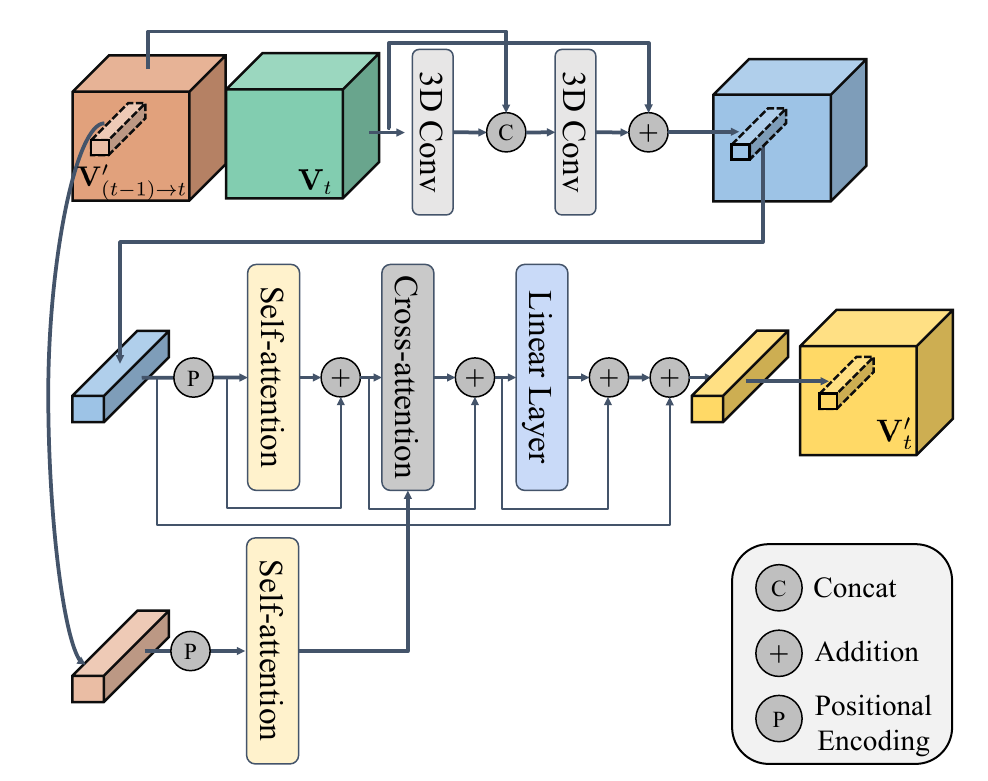}    
    \caption{    
    Our ray fusion module.
    }    
	\label{fig:fusion_module}
\end{wrapfigure}

Instead, we propose the \emph{ray-wise fusion scheme} that calculates attention for only extracted ray-wise features of the aligned cost volumes $\mathbf{V}_{t}$ and $\mathbf{{V}}'_{(t-1)\rightarrow t}$ (\Figs{concept} and \ref{fig:fusion_module}). For an arbitrary pixel position $(h,w)$, we can obtain a ray-wise feature $\mathbf{F}_t =\mathbf{V}(:,:,h,w) \in \mathbb{R}^{D \times C}$ from a cost volume, where each row of the matrix $\mathbf{F}_t$ indicates a $\textrm{C}$-dimensional feature vector for a certain depth plane hypothesis. We then regard $\mathbf{F}_t$ as $\textrm{D}$ tokens, where each token is a $\textrm{C}$-dimensional feature, and apply the attention mechanism to fuse two ray-wise features $\mathbf{F}_t=\mathbf{V}_{t}(:,:,h,w)$ and $\mathbf{F}_{t-1}=\mathbf{V}'_{(t-1)\rightarrow t}(:,:,h,w)$.


A straightforward option for the fusion is to compute cross-attention between $\mathbf{F}_t$ and $\mathbf{F}_{t-1}$. However, the cross-attention does not consider the intrinsic properties of individual ray-wise features. Inspired by the stereo matching approach~\cite{won2020end} that utilizes entropy of a probability distribution as an uncertainty prior, we expect the network to consider the intrinsic characteristics of each ray-wise feature.
In addition, this independent fusion does not consider spatially adjacent features within cost volumes. 

To compensate for such deficiency, we apply two 3D convolutional layers to the volumes before applying our fusion scheme, as shown in Figure~\ref{fig:fusion_module}. We then compute the self-attention
$\mathbf{SA}_t = Attn(\textbf{F}_t, \textbf{F}_t, \textbf{F}_t)$\footnote{denoted as $Attn(\mathbf{Q},\mathbf{K},\mathbf{V}) = \bigl(softmax(\frac{\mathbf{Q}\mathbf{W}_Q(\mathbf{K}\mathbf{W}_K)^\mathsf{T}}{\sqrt{d}} )\mathbf{V}\mathbf{W}_V \bigl) \mathbf{W}_O$, where \{$\mathbf{W}_Q$, $\mathbf{W}_K$, $\mathbf{W}_V$, $\mathbf{W}_O$\} are learnable linear projection parameters.} and compute $\mathbf{SA}_{t-1}$ similarly. Finally, the fused feature is calculated using cross-attention $\mathbf{CA}_t = Attn(\mathbf{SA}_t,$ $\mathbf{SA}_{t-1},$ $\mathbf{SA}_{t-1})$. To inject information about relative positions among $\textrm{D}$ tokens in $\mathbf{F}$, we add the sinusoidal positional encodings~\cite{vaswani2017attention} of depth plane indices before the fusion.

We repeat the process for all ray-wise feature pairs to make the fused cost volume ($\mathbf{V}'_{t}$) as depicted in \Fig{fusion_module}). Note that the proposed method needs only $\textrm{D}^2\textrm{HW}$ entries for constructing attention maps, and it is much more memory-efficient than na\"ive approach consuming $\textrm{D}^2\textrm{H}^2\textrm{W}^2$ entries. 

\subsection{Depth Regression} \label{sec:regress}

To regress a completed depth map from the fused cost volume $\mathbf{V}'_{t} \in \mathbb{R}^{D \times C \times H \times W}$, we firstly convert the fused cost volume to an unnormalized probability volume $\mathbf{P}'_t$ ($\textrm{D}\times \textrm{H}\times \textrm{W}$) via a single 3D convolutional layer and per-plane pixel shuffle~\cite{kam2022costdcnet}.
Then, we apply the softmax operator $\sigma(\cdot)$~\cite{kendall2017end} to regress a completed depth $\mathbf{D}_t$ as follows~\cite{gu2020cascade,im2019dpsnet}:
\begin{gather}\label{eq:regress}
\mathbf{D}_t(h,w) = \sum_{i=1}^{D}{d_i} \times \mathbf{p}^i_{h,w}, \quad \mathbf{p}_{h,w}=\mathbf{P}_t(:,h,w)=\sigma(\mathbf{P}'_t(:,h,w)), 
\end{gather}
where $d_i$ is the pre-defined depth value of the $i$-th hypothesis depth plane, $(h,w)$ is an image pixel position, $D$ is the number of hypothesis depth planes, $\mathbf{P}_t$ is a probability volume, and $\mathbf{p}_{h,w}$ is a $D$-dimensional probability vector for depth planes at $(h,w)$.

To further refine the regressed depth $\mathbf{D}_t$ on the image domain, we adapt non-local spatial propagation networks (NLSPN)~\cite{park2020non} with minor modification. NLSPN takes as the input an affinity map, a confidence map, and a depth to be refined.
In our case, the regressed depth is accurate enough, so we utilize shallow 2D convolutional networks for estimating an affinity map, and we directly compute a confidence map from $\mathbf{P}_t(d,h,w)$.
For more details, we refer the readers to the supplementary document.

\subsection{Loss Function}
Our framework is fully differentiable, and it can be trained in an end-to-end manner. We use the $L_1$ depth loss and a cross-entropy loss for probability volume supervision (\Fig{overview}). 
$L_1$ depth regression loss is defined as follows:
\begin{equation}
\mathcal{L}_{L1} = \frac{1}{|\mathbb{P}|} \sum_{(h,w) \in \mathbb{P} } \left | \mathbf{D}_t(h,w) - \mathbf{D}_{gt}(h,w) \right |,
\end{equation}
where $\mathbb{P}$ is the set of sparse GT depth pixels, $\mathbf{D}_t$ and $\mathbf{D}_{gt}$ are the completed depth and a ground truth depth. 
The cross-entropy loss is defined as follows:
\begin{equation}
\mathcal{L}_{CE} = \frac{1}{|\mathbb{P}|} \sum_{(h,w) \in \mathbb{P} } -\mathbf{p}_{gt}^{\mathsf{T}}\log\mathbf{p} ,
\end{equation}
where $\mathbf{p}_{gt}$ is a ground truth probability vector over the hypothesis depth plane, and $\mathbf{p}$ is a predicted probability vector obtained from $\mathbf{P}_t(:,h,w)$ (\Eq{regress}). 

We found that using hard labels (one-hot vector) for $\mathbf{p}_{gt}$ is not effective for the performance, similar to observations in \cite{kendall2017end}. We instead make soft labels from ground truth depths using the idea of Nuanes et al.~\cite{Nuanes_2021_CVPR}. We find two nearest depth planes for a given ground truth depth and then compute normalized weights for respective planes. This computation results in a probability distribution vector where all elements are zero except for two elements containing the respective normalized values.

Finally, our total loss $\mathcal{L}_{total}$ is defined as follows:
\begin{equation}
\mathcal{L}_{total} = \mathcal{L}_{L1} + \mathcal{L}_{CE}.
\end{equation}

\section{Experiments}
\subsection{Implementation Details}

We implement our framework using PyTorch~\cite{paszke2019pytorch, Falcon_PyTorch_Lightning_2019, rw2019timm}. We train our model with a batch size of four on three NVIDIA GeForce RTX 3090 GPUs. We employ the AdamW optimizer~\cite{loshchilov2017decoupled} with a weight decay of $0.0001$ and an initial learning rate of $0.001$. The learning rate is reduced by a factor of 0.5 at a predefined epoch schedule. The average inference time of our model on the ScanNetv2 test set is 77ms. The number of hypothesis depth planes is set to $16$. 
A cost volume as a four-dimensional volume requires a high memory footprint. In practice, we utilize downscaled images with a factor of 4.
The supplementary document provides additional visual results and detailed information, including network architecture, error metrics, GPU memory consumption, and inference times.

\subsection{Datasets}
\label{sec:datasets}
We use the following indoor/outdoor datasets to demonstrate our approach.

\textbf{ScanNetV2}~\cite{dai2017scannet} dataset is a large-scale RGB-D dataset for indoor scenes, comprising 1,201 training, 312 validation, and 100 testing scenes ($\approx$211,000 frames) captured with a handheld RGB-D sensor. This dataset includes accurate camera calibration parameters and ground truth depth images. In this case, we randomly obtain 300 depth samples from a ground truth depth image to create input sparse depth images. We set $d_{\textrm{min}}=10^{-3}m$ and $d_{\textrm{max}}=10m$. We randomly cropped images to $512\times384$ pixels during training. The learning rate schedule is $\{10,15,20,25\}$ over 30 epochs. We test depths less than 10m for performance evaluation.

\textbf{VOID}~\cite{wong2020unsupervised} dataset contains synchronized $640\times480$ RGB images and sparse depth maps of indoor (e.g., laboratories and classrooms) and outdoor (gardens) scenes.
The depth maps contain about 150, 500, or 1500 sparse depth samples (corresponding to 0.05\%, 0.15\%, and 0.5\%) for each scene. Depth samples are obtained from a set of image features tracked by XIVO~\cite{fei2019geo},
a visual-inertial odometry system. The dense ground-truth depth maps are acquired by active stereo.
The VOID dataset contains 56 sequences
with challenging camera motions. Among the 56 sequences, 48 sequences ($\approx$45,000 frames) are
designated for training, and eight sequences (800 frames) are assigned for testing. We set $d_{\textrm{min}}=10^{-3}m$ and $d_{\textrm{max}}=6m$. During training, randomly cropped images of $512\times384$ pixels are used, and the learning rate schedule is $\{30,40,50,60,70\}$ over 80 epochs.
We follow the evaluation protocol of \cite{wong2020unsupervised} and evaluate depths within [0.2, 5.0]m.

\textbf{KITTI}~\cite{Uhrig2017THREEDV} depth completion (DC) benchmark dataset contains about 86,000  $1242\times 375$ RGB-D pairs that capture diverse road scenes. The sparse depth samples are obtained using a Velodyne LiDAR sensor, and it accounts for approximately 5\% of the image space. As the 1,000 test set frames of KITTI are not captured sequentially, our temporal fusion module cannot be applied. Thus, we evaluate our method on scenes in the validation set with sequential frames. We set $d_{\textrm{min}}=10^{-1}m$ and $d_{\textrm{max}}=90m$. During training, we use randomly cropped images of $1216\times240$ pixels, and the learning rate decay schedule is $\{30,40,50,60,70\}$ over the total 100 epochs.

\begin{table}[t]
\caption{Quantitative comparison on the ScanNetV2~\cite{dai2017scannet} test set. `SPN' and `S' denote our depth refinement module and single-view setting. `R' means that GT depths are obtained by rendering GT meshes. The unit for all metrics, except for \textbf{iMAE} (1/m) and \textbf{iRMSE} (1/m), is meter.}

 \label{tbl:scannet_dataset}
\centering
\resizebox{0.99\linewidth}{!}{
\begin{tabular}{crrrrr|rrrrrr}
\toprule
\multirow{2}{*}{\textbf{Method}} & \multirow{2}{*}{\textbf{\#Param.}} & \multicolumn{4}{c}{\textbf{Depth Error}} & \multicolumn{6}{c}{\textbf{3D Reconstruction Error}}\\
\cline{3-12} 
   & & \textbf{MAE}$\downarrow$ & \textbf{RMSE}$\downarrow$ & \textbf{iMAE}$\downarrow$ & \textbf{iRMSE}$\downarrow$  & \textbf{Acc}$\downarrow$ & \textbf{Compl}$\downarrow$ & \textbf{Chamfer}$\downarrow$ & \textbf{Prec}$\uparrow$ & \textbf{Recall}$\uparrow$ & \textbf{F-score}$\uparrow$ \\
\hline
SimpleRecon~\cite{sayed2022simplerecon} &	49.1M&	{0.0887}	&{0.1448}	&{0.0315}	&{0.0499}  &0.0681 &0.0557 &0.0619 &0.6694 &0.6494 &0.6572\\ 
ComplFormer~\cite{zhang2023completionformer}	& 83.5M	&0.0276&	0.0868&	{0.0091}	&0.0295 & 0.0442   &{0.0223}	        & 0.0332  &0.8596  &{0.9152}    &0.8838\\
NLSPN~\cite{park2020non}& {25.8M} &{0.0266} &	{0.0859}&	{0.0086}&{0.0290} & {0.0412}           & 0.0223	        & {0.0318}   & {0.8642}  & 0.9147          &{0.8862}\\
CostDCNet~\cite{kam2022costdcnet}	&
1.8M& 
0.0244 &  0.0759 & 
0.0086 &
0.0255 &
{0.0339}   &
{\color{silver}\faMedal}{0.0204}   &
{0.0272}   &
{\color{silver}\faMedal}{0.8778}  &
{0.9269}  &
{0.8998}\\
\hline
Ours w\slash o SPN+S 	&{\color{golden}\faMedal}{1.11M}	&
0.0208&
{\color{silver}\faMedal}0.0681 & {\color{silver}\faMedal}0.0068&
0.0230 & -& -& -& -& -&- \\

Ours  w\slash o SPN &{\color{golden}\faMedal}{1.11M}	&{\color{golden}\faMedal}{0.0159}	&{\color{golden}\faMedal}{0.0553}	&{\color{golden}\faMedal}{0.0053}	&{\color{golden}\faMedal}{0.0192} &{\color{golden}\faMedal}{0.0294}   &{\color{golden}\faMedal}{0.0181}   &{\color{golden}\faMedal}{0.0237}  &{\color{golden}\faMedal}{0.8908}   &{\color{golden}\faMedal}{0.9467}  &{\color{golden}\faMedal}{0.9163}\\

Ours  	&{\color{silver}\faMedal}{1.15M}	&{\color{silver}\faMedal}{0.0160}	&{\color{silver}\faMedal}{0.0554}	&{\color{golden}\faMedal}{0.0053}	&{\color{silver}\faMedal}{0.0193} & {\color{silver}\faMedal}{0.0295}   &{\color{golden}\faMedal}{0.0181}   &{\color{silver}\faMedal}{0.0238}  &{\color{golden}\faMedal}{0.8908}   &{\color{silver}\faMedal}{0.9465}  &{\color{silver}\faMedal}{0.9161}\\ 

\hline
\hline
DeepSmooth+R  	&20.4M	&0.043  &	0.142	& - &- & -   &-   &-  &-   &- & -\\ 

Ours+S+R  	&{\color{golden}\faMedal}{1.15M}	&{\color{golden}\faMedal}{0.036}	&{\color{golden}\faMedal}{0.114} &0.0199 &0.0624 & -   &-   &-  &-   &-  & -\\ 

\bottomrule
\end{tabular}
}

\end{table}

\begin{table}[t]
\caption{Quantitative comparison on VOID~\cite{wong2020unsupervised} test set. `SPN' and `S' denote our depth refinement module and single-view setting. }

 \label{tbl:void_1500}
\centering
\resizebox{0.65\linewidth}{!}{
\begin{tabular}{@{}c@{}r|c|c|rrrr@{}}
\toprule
\multirow{2}{*}{\textbf{Method}} 
     & \multirow{2}{*}{\textbf{\#Param.}} & \multicolumn{2}{c}{\textbf{Density}} & {\textbf{MAE}$\downarrow$} &  {\textbf{RMSE}$\downarrow$} &  {\textbf{iMAE}$\downarrow$} & {\textbf{iRMSE}$\downarrow$}  \\
     \cline{3-4}
 & & Training&Testing & (mm) & (mm)& (1/km)& (1/km) \\
\hline 
SS-S2D~\cite{ma2018self}&	27.8M&	\multirow{11}{*}{0.50\%}& \multirow{11}{*}{0.50\%} &178.85&	243.84&	80.12&	107.69\\
DDP~\cite{yang2019dense}&	18.8M&   &    &151.86&	222.36&	74.59&	112.36\\
VOICED~\cite{wong2020unsupervised}&	9.7M&   &     &	85.05	&169.79&	48.92&	104.02\\
ScaffNet~\cite{wong2021learning}&	7.8M&    &   &	59.53&	119.14&	35.72	&68.36\\
MSG-CHN~\cite{li2020multi}&	{\color{golden}\faMedal}{0.36M}&         &        &	43.57&	109.94&	23.44&	52.09\\
KBNet~\cite{wong2021unsupervised}&	6.9M&         &        &	39.80&	95.86&	21.16	&49.72\\
PENet~\cite{hu2020PENet}&	132.0M&         &        &	34.61&	82.01	&18.89	&40.36\\
Mondi~\cite{liu2022monitored}&	5.3M&         &        &	29.67&	79.78&	14.84	&37.88\\
NLSPN~\cite{park2020non}&	25.8M&         &        	&{26.74}	&{79.12}&	{12.70}&	{33.88}\\
ComplFormer~\cite{zhang2023completionformer}&	83.5M&         &        	&49.61	&141.40&	21.08&	51.53\\
LRRU~\cite{LRRU_ICCV_2023}& 21.0M & & & 47.20   & 118.00  &	 22.00  &	48.30 \\
CostDCNet~\cite{kam2022costdcnet}&	{1.8M}&         &        &	{25.84}&	
{\color{bronze}\faMedal}
{76.28}&
{\color{bronze}\faMedal}
{12.19}
&{32.13}\\
\hline
 Ours w\slash o SPN+S&	 {\color{silver}\faMedal}1.11M &\multirow{3}{*}{0.50\%}& \multirow{3}{*}{0.50\%}&	{\color{bronze}\faMedal}25.53&	{\color{silver}\faMedal}68.83 & 12.37&{\color{bronze}\faMedal}31.52 \\
 Ours w\slash o SPN&	{\color{silver}\faMedal}1.11M &           &           & {\color{silver}\faMedal}{24.57}& {\color{golden}\faMedal}{65.46}	&{\color{silver}\faMedal}{12.03}& {\color{silver}\faMedal}{30.26}\\
Ours&	{\color{bronze}\faMedal}{1.15M}&           &          	&{\color{golden}\faMedal}{24.51}	&{\color{golden}\faMedal}{65.46}	&{\color{golden}\faMedal}{11.98}	&{\color{golden}\faMedal}{30.20} \\

\hline
\hline
\multirow{2}{*}{LRRU~\cite{LRRU_ICCV_2023}}	&\multirow{2}{*}{21.0M}	&  \multirow{2}{*}{0.50\%} & 0.15\%& 115.30   & 262.60  &	44.70 &	86.30 \\
 	& 	&  & 0.05\%&207.80   & 409.90&78.20 &127.30 \\
\hline 
\multirow{2}{*}{ComplFormer~\cite{zhang2023completionformer}}	&\multirow{2}{*}{83.5M}	&  \multirow{2}{*}{0.50\%} & 0.15\%&228.49&	449.51&	62.71&	119.92\\
 	& 	&  & 0.05\%&395.42	&639.05	&107.52&	174.05\\
\hline 
\multirow{2}{*}{NLSPN~\cite{park2020non}}	&\multirow{2}{*}{25.8M}&  \multirow{2}{*}{0.50\%}& 0.15\%	&{65.91}	&{160.76}&	{27.79}& {63.26}\\
 	& 	&    & 0.05\%&{118.19}	&{245.41}&	{52.57}	&{99.36} \\
\hline 
\multirow{2}{*}{Mondi~\cite{liu2022monitored}}&	\multirow{2}{*}{{\color{bronze}\faMedal}{5.3M}}	&  \multirow{2}{*}{0.50\%}& 0.15\% &{61.37}&	{146.57}	&{27.96}&	{64.36}\\
 &			&   & 0.05\% & {104.97}&	{225.60}	&{48.44}& {96.79} \\
 \hline 

\multirow{2}{*}{Ours w\slash o SPN+S}	&\multirow{2}{*}{{\color{golden}\faMedal}{1.11M}}& \multirow{2}{*}{0.50\%} & 0.15\% & {\color{bronze}\faMedal}52.80&	{\color{bronze}\faMedal}121.65 & {\color{bronze}\faMedal}24.80 &{\color{bronze}\faMedal}56.83 \\
 	&     &    & 0.05\%&	{\color{bronze}\faMedal}84.45&	{\color{bronze}\faMedal}174.46 & {\color{bronze}\faMedal}41.75&{\color{bronze}\faMedal}83.64\\
\hdashline 
\multirow{2}{*}{Ours w\slash o SPN}	& \multirow{2}{*}{{\color{golden}\faMedal}{1.11M}} & \multirow{2}{*}{0.50\%}& 0.15\% & {\color{silver}\faMedal}{48.75}&	{\color{silver}\faMedal}{110.68}	&{\color{silver}\faMedal}{23.17}&{\color{silver}\faMedal}{52.43}  \\
 	& &  & 0.05\%&	{\color{silver}\faMedal}{78.65}& {\color{silver}\faMedal}{162.32} &{\color{silver}\faMedal}{39.22} & {\color{silver}\faMedal}{77.65} \\
\hdashline 
\multirow{2}{*}{Ours}	& \multirow{2}{*}{{\color{silver}\faMedal}{1.15M}} &\multirow{2}{*}{0.50\%} & 0.15\% & {\color{golden}\faMedal}{48.67}&	{\color{golden}\faMedal}{110.56}&	{\color{golden}\faMedal}{23.10}& {\color{golden}\faMedal}{52.35} \\
 	& &  & 0.05\%&	{\color{golden}\faMedal}{78.55}	&{\color{golden}\faMedal}{162.17}	&{\color{golden}\faMedal}{39.15}	&{\color{golden}\faMedal}{77.55}\\

\bottomrule
\end{tabular}
}

\end{table}

\begin{table}[t]
\caption{ Quantitative comparison on the KITTI~\cite{Uhrig2017THREEDV} validation set. `SPN' and `S' denote our depth refinement module and single-view setting.
}

 \label{tbl:kitti_dataset}
\centering
\resizebox{0.60\linewidth}{!}{
\begin{tabular}{@{}crrrrr@{}}
\toprule
    
  \multirow{2}{*}{\textbf{Method}} & \multirow{2}{*}{\textbf{\#Param.}} & \textbf{MAE}$\downarrow$ & \textbf{RMSE}$\downarrow$ & \textbf{iMAE}$\downarrow$ & \textbf{iRMSE}$\downarrow$ \\
  & & (mm) & (mm) & (1/km) & (1/km)\\
\hline 
SS-S2D~\cite{ma2018self}&	27.8M&	269.20&	878.50&	1.34&	3.25 \\
Depth-normal~\cite{xu2019depth}&	29.0M&	236.67&	811.07&	1.11&	2.45 \\
MSG-CHN~\cite{li2020multi}&	{\color{golden}\faMedal}{0.36M}&	227.94&	821.94&	0.98&	2.47\\
Mondi~\cite{liu2022monitored}&	5.3M&	218.22&	815.16&	0.91&	2.18\\
Uber-FuseNet~\cite{chen2019learning}&	1.9M&	217.00&	785.00&	1.08&	2.36\\
DeepLidar~\cite{qiu2019deeplidar}&	53.4M&	215.38&	{\color{golden}\faMedal}{687.00}&	1.10&	2.51\\
DC-3co~\cite{imran2019depth}&	27.0M&	215.04&	1011.30&	0.94&	2.50\\
3DepthNet~\cite{xiang20203ddepthnet}&	     -&	208.96&	{\color{silver}\faMedal}{693.23}&	0.98&	2.37\\
PENet~\cite{hu2020PENet}&	131.5M&	208.81&	753.75&	0.91&	2.16\\
NLSPN~\cite{park2020non}&	25.8M&	198.64&	771.80&	0.83&	2.03\\
CompletionFormer~\cite{zhang2023completionformer} &	83.5M&	198.63&	748.07&	0.85&	{2.00}\\
TWISE~\cite{imran2021depth}&	{1.5M}&	193.40&	879.40&	{0.81}&	2.19\\
DySPN~\cite{lin2022dynamic}&	26.3M&	{192.50}&	745.80&	-&	-\\
LRRU~\cite{LRRU_ICCV_2023}&	21.0M&	{\color{bronze}\faMedal}188.80&	729.50&	{0.80}&	{\color{bronze}\faMedal}{1.90}\\
\hline
Ours w\slash o SPN+S&	 {\color{silver}\faMedal}1.11M&	{193.05}&	{793.03} & {\color{bronze}\faMedal}0.76&{\color{silver}\faMedal}1.74\\
Ours w\slash o SPN&	{\color{silver}\faMedal}1.11M& {\color{silver}\faMedal}{182.65}&	{726.95}	&{\color{silver}\faMedal}{0.74}& {\color{golden}\faMedal}{1.68}\\
Ours&	{\color{bronze}\faMedal}1.15M&	{\color{golden}\faMedal}{176.23}
&	{\color{bronze}\faMedal}{720.63}	&{\color{golden}\faMedal}{0.72}&	{\color{golden}\faMedal}{1.68}\\

\toprule
\end{tabular}
}

\end{table}
\begin{figure}[t]
	\centering	
\includegraphics[width=0.99\textwidth]{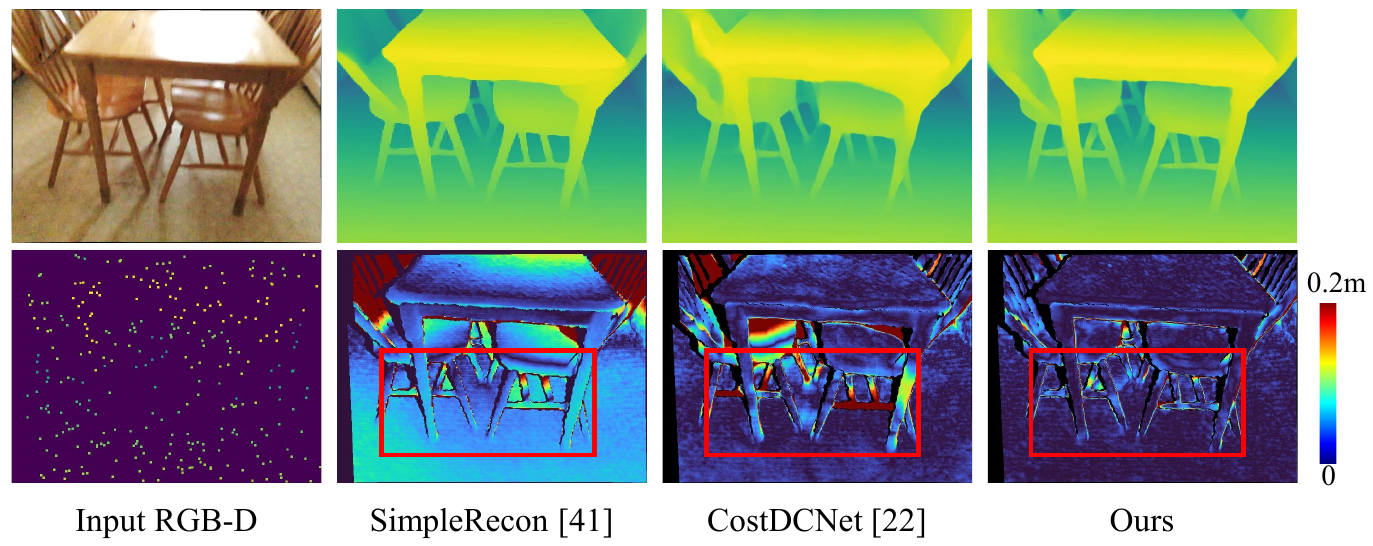}\\
        
	\caption{Visual comparison of completed depths (top) on the ScanNetV2 test set. Error maps of completed depths are also presented (bottom).
	}
	\label{fig:scannet_depth}
 
\end{figure}


\begin{figure}[t]
	\centering	
\includegraphics[width=0.99\linewidth]{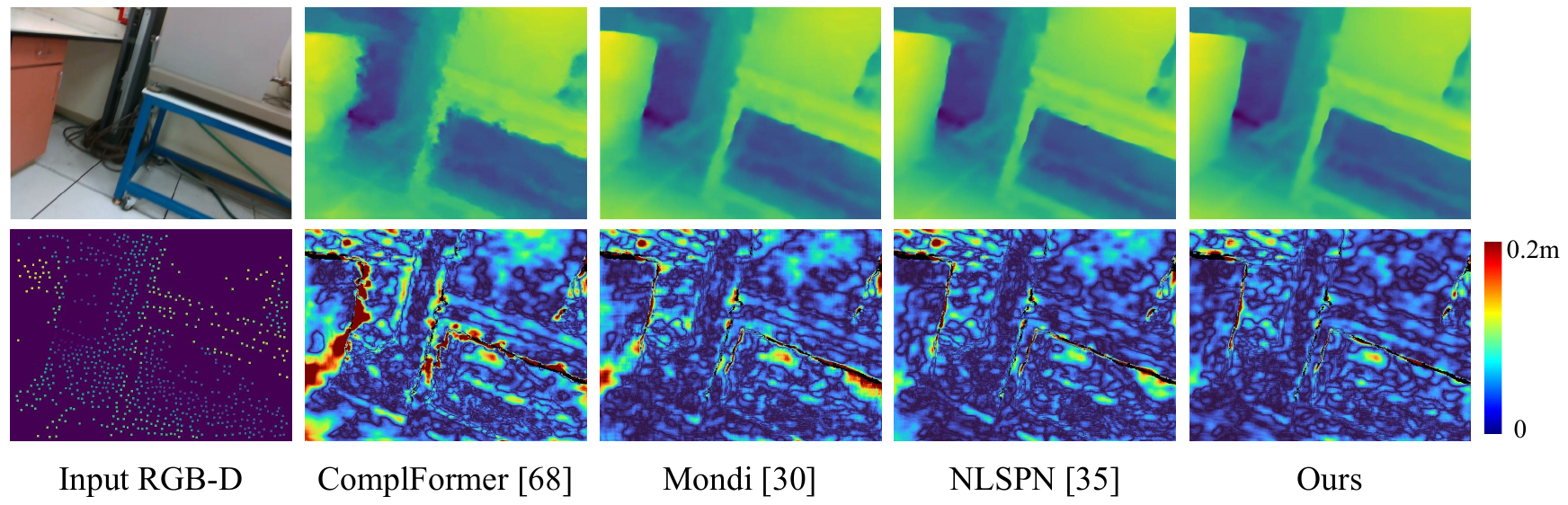}\\

 \caption{Visual comparison of our method with state-of-the-art depth completion methods~\cite{zhang2023completionformer,liu2022monitored,park2020non} on the VOID test set (0.5\% sparsity)~\cite{wong2020unsupervised}. Completed depths (top) and error maps (bottom) are presented.  
 }
 
	\label{fig:void_depths}
\end{figure}

\begin{figure}[t]
	\centering
\includegraphics[width=0.99\linewidth]{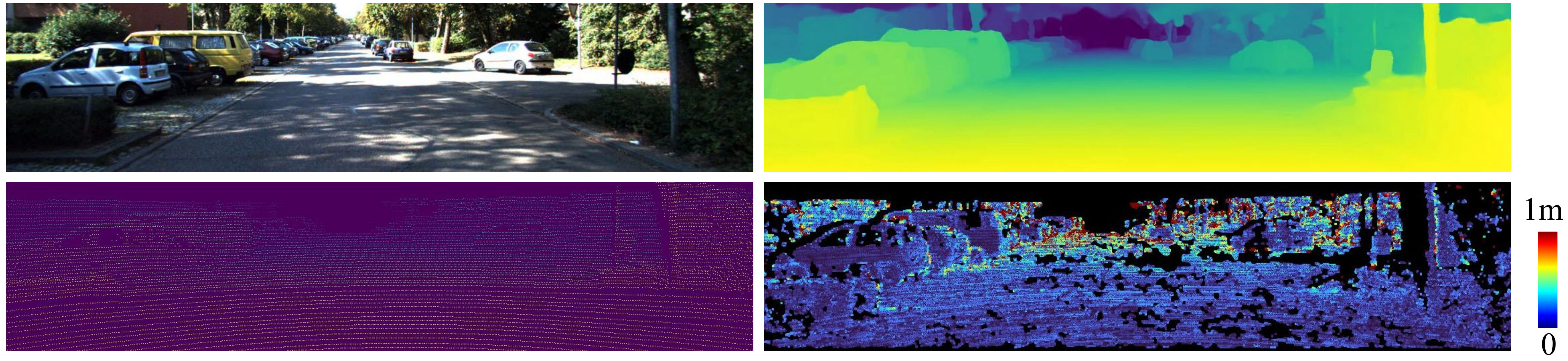}\\
    
	\caption{Our completion result on the KITTI depth completion benchmark. Please refer to the supplementary document for a comparison with other approaches.}
 
	\label{fig:kitti_depths}	
\end{figure}

\subsection{Comparison}

We compare our method with state-of-the-art (SOTA) single-view and video depth completion models qualitatively and quantitatively. For extensive comparison, we also present a comparison with a recent multiview stereo method~\cite{sayed2022simplerecon} using RGB videos. 

In the ScanNetV2 experiment, we choose recent single-view sparse depth completion approaches (NLSPN~\cite{park2020non}, CostDCNet~\cite{kam2022costdcnet}, CompletionFormer~\cite{zhang2023completionformer}), a video depth completion method (DeepSmooth~\cite{Krishna_2023_CVPR}), and a multiview stereo method (SimpleRecon~\cite{sayed2022simplerecon}). We used the pretrained model of SimpleRecon provided by the authors, and we trained other models from scratch using the official source codes. Since DeepSmooth is a framework designed for semi-dense depth completion and its official source code is not available, we trained our framework under the same conditions as DeepSmooth for comparison. 
As a result, our method shows the best performance in both depth and 3D error metrics (\Tbl{scannet_dataset} and \Fig{scannet_depth}). While SimpleRecon often produces visually pleasing depth maps, its estimated depth values result in inferior quantitative performance because it does not utilize sparse depth samples, which serve as a strong prior.

We observe similar trends in the VOID test set, even if the depth samples provided by the VOID dataset are not from range-based sensors. Our model outperforms the state-of-the-art approaches~\cite{ma2018self,yang2019dense,wong2020unsupervised,wong2021learning,li2020multi,wong2021unsupervised,hu2020PENet,liu2022monitored,park2020non,zhang2023completionformer,kam2022costdcnet} in most error metrics in this VOID dataset. Compared to CostDCNet, the second-best model, our approach shows better performance with the smaller network parameters (\Tbl{void_1500} and \Fig{void_depths}).

We achieved the best score among the baseline approaches~\cite{ma2018self,xu2019depth,li2020multi,liu2022monitored,chen2019learning, qiu2019deeplidar,imran2019depth,xiang20203ddepthnet,hu2020PENet,park2020non,zhang2023completionformer,imran2021depth,lin2022dynamic,LRRU_ICCV_2023} in the validation set of the KITTI depth completion benchmark in most of the error metrics. Note that modern approaches such as DySPN~\cite{lin2022dynamic}, NLSPN~\cite{park2020non}, CompletionFormer~\cite{zhang2023completionformer}, and LRRU~\cite{LRRU_ICCV_2023} have much larger number of network parameters than ours to achieve competitive performance (\Tbl{kitti_dataset} and \Fig{kitti_depths}). We believe that our ray-based cost fusion scheme using self-/cross-attention is effective on various datasets.

\subsection{Analysis}

\noindent\textbf{Effect of ray-based cost volume fusion.} 
Our ray-based fusion improves the performance in most of the metrics with a small number of additional parameters (0.08M) (\Tbl{kitti_ablation} (\romannumeral3, \romannumeral7)). To compare the local cost volume fusion and our ray-based fusion, we adopt a convolutional variant of gated recurrent unit (GRU) used in~\cite{sun2021neucon} to fuse two aligned cost volumes. As a result, while the local fusion achieves slight performance improvement in the RMSE metric, its MAE metric rather deteriorates (\Tbl{kitti_ablation} (\romannumeral5)).

\noindent\textbf{Effect of cross entropy loss.} 
We observed that using $\mathcal{L}_{CE}$ loss for direct  distribution supervision contributes to stable learning. It also results in a balanced performance between mean absolute difference (MAE) and root mean square error (RMSE). This is a different observation to existing depth completion works~\cite{park2020non,lin2022dynamic,zhang2023completionformer} that combine $L_1$ and $L_2$ depth loss terms (\Tbl{kitti_ablation} (\romannumeral1) and (\romannumeral2)). A combination of three losses ($L_1$ + $L_2$ + CE loss term) does not lead to performance gain compared to using only the CE loss term (\Tbl{kitti_ablation} (\romannumeral2), (\romannumeral4)).

\noindent\textbf{Single RGB-D image as input.}
To evaluate the performance of our framework given a single RGB-D image as the input, we disable the cross-attention part of our fusion module and use only the self-attention part. Inevitably, the performance degrades in this single-view setting as it cannot leverage information from previous RGB-D frames. However, it is noteworthy that even in the single-view scenario, our framework is competitive to state-of-the-art depth completion methods~\cite{zhang2023completionformer,lin2022dynamic,LRRU_ICCV_2023,liu2022monitored,park2020non,kam2022costdcnet} on three different types of datasets (VOID, KITTI, and ScanNet) (\Tbls{kitti_dataset}, \ref{tbl:void_1500}, \ref{tbl:scannet_dataset}).

\noindent\textbf{Effect of depth refinement module.} 
Our depth refinement module, which has small network parameters (0.04M), improves the MAE metric on the KITTI dataset. However, we observed that the impact of the NLSPN on performance in the VOID dataset is relatively marginal, and the addition of NLSPN even leads to a slight decrease in performance on the ScanNetV2 dataset (\Tbls{kitti_dataset}, \ref{tbl:void_1500}, \ref{tbl:scannet_dataset}).      

\noindent\textbf{Robustness to various depth sparsity.} 
To assess the sparsity-agnostic capability of our model and other approaches, we train the models on the VOID dataset with 0.5\% depth sparsity and evaluate them on the VOID test set having 0.15\% and 0.05\% sparsity. As shown in \Tbl{void_1500}, our approach is less affected by the changed depth sparsity compared with other approaches~\cite{liu2022monitored,zhang2023completionformer,park2020non}, despite using smaller network parameters.

\noindent\textbf{Cross-dataset generalization.} 
To evaluate the cross-dataset generalization ability, we train our model and baseline approaches (NLSPN~\cite{park2020non} and CompletionFormer~\cite{zhang2023completionformer}) on the ScanNetv2 training set and evaluate them on the VOID test set with $0.5\%$ density. While CompletionFormer performs slightly worse than NLSPN on the ScanNetV2 test set, it demonstrates better generalization ability on the VOID test set (see \Tbls{scannet_dataset} and \ref{tbl:cross_dataset}). Our method exhibits better generalizability than other methods. We speculate that our approach, which directly works with ray-wise cost slice, is a more generic approach across different dataset domains.

\begin{table}[t]
\caption{ Ablation study on the KITTI validation set. `A' denotes our framework without the proposed ray-based fusion `B1' and depth refinement `C'. `B2' denotes the convolutional GRU-based cost volume fusion~\cite{sun2021neucon}. `$L_1$', `$L_2$', and `$L_{CE}$' denote $L_1$ and $L_2$ depth losses, and cross entropy loss, respectively. We use 20\% of the KITTI training set for this experiment.}

 \label{tbl:kitti_ablation}
\centering
\resizebox{0.70\linewidth}{!}{

\begin{tabular}{@{}rlcrrrr@{}}
\toprule
 & \textbf{Network / Loss} & \textbf{\#Param.} & \textbf{MAE}$\downarrow$ & \textbf{RMSE}$\downarrow$ & \textbf{iMAE}$\downarrow$ & \textbf{iRMSE}$\downarrow$  \\
\hline 
\romannumeral 1 & A / $L_1$ & 1.03M & {203.73} & 855.07 & {\color{silver}\faMedal}{0.79} & {1.85} \\
\romannumeral 2 & A / $L_{CE}$ & 1.03M & 223.59 & 812.49 & 0.91 & 1.98 \\
\romannumeral 3 & A / $L_{CE}$+$L_1$ & 1.03M & {203.05} & 834.09 & {\color{silver}\faMedal}{0.79} & {1.83} \\
\romannumeral 4 & A / $L_{CE}$+$L_1$+$L_2$  & 1.03M & 220.22 & 817.94 & 0.89 & 1.98 \\
\romannumeral 5 & A+B2 / $L_{CE}$ & 1.09M & 228.85 & {799.03} & 1.03 & 2.04 \\
\romannumeral 6 & A+B1 / $L_{CE}$ & 1.11M & 215.18 & {\color{silver}\faMedal}{774.10} & {0.88} & 1.87 \\
\romannumeral 7 & A+B1 / $L_{CE}$+$L_1$ & 1.11M & {\color{silver}\faMedal}{198.51} & {777.37} & {0.82} & {\color{silver}\faMedal}{1.82}\\
\romannumeral 8 & A+B1+C / $L_{CE}$, SPN $L_1$ & 1.15M & {\color{golden}\faMedal}{188.88} & {\color{golden}\faMedal}{768.11} & {\color{golden}\faMedal}{0.77} & {\color{golden}\faMedal}{1.78} \\

\toprule
\end{tabular}
}

\end{table}

\begin{table}[t]
\caption{ Quantitative comparison for cross-dataset generalization ability. All models are trained on the ScanNetv2 training set and tested on the VOID test set of 0.5\% depth density. }

 \label{tbl:cross_dataset}
\centering

\resizebox{0.60\linewidth}{!}{
\begin{tabular}{@{}cccccc@{}}
\toprule
 \textbf{Method} & \textbf{Params} & \textbf{MAE}$\downarrow$ & \textbf{RMSE}$\downarrow$ & \textbf{iMAE}$\downarrow$ & \textbf{iRMSE}$\downarrow$  \\
\hline 
NLSPN~\cite{park2020non}	&25.8M	&158.60 & 571.80  &	22.00&	57.50\\
CompletionFormer~\cite{zhang2023completionformer} &83.5M & 65.90 &	190.01&	 20.66&	49.83\\
Ours	&{\color{golden}\faMedal}{1.15M} & {\color{golden}\faMedal}{29.08}&	{\color{golden}\faMedal}{78.34}&	{\color{golden}\faMedal}{12.79}&	{\color{golden}\faMedal}{31.93} \\
\toprule
\end{tabular}
}

\end{table}

\section{Conclusion}
In this paper, we proposed a learning-based depth completion framework that effectively utilizes temporal information from an RGB-D video. We introduced the ray-based cost volume fusion scheme that leverages the attention mechanism. The fusion module effectively fuses cost volume predictions over time to infer a more accurate cost volume which is used for completed depth regression subsequently. We demonstrate that our framework, \textit{\Ours{}}, consistently beats or rivals state-of-the-art (SOTA) depth completion methods on diverse indoor and outdoor datasets, despite utilizing significantly fewer network parameters.

\noindent\textbf{Limitation and future work.}
Our framework, relying on cost volumes with 3D convolutions and computing attention maps between them, suffers from a high memory footprint. Additionally, persistent poor depth predictions over time pose challenges that our method cannot fully resolve. Future work could involve designing a more computationally efficient network architecture~\cite{sayed2022simplerecon} that eliminates fully 3D convolutions.

\clearpage

\section*{Acknowledgements}
This work was supported by the NRF grant (RS-2023-00280400) and IITP grants (ICT Research Center, RS-2024-00437866; RS-2023-00227993; AI Innovation Hub, RS-2021-II212068; AI Graduate School Programs at POSTECH and SNU, RS-2019-II191906 and RS-2021-II211343) funded by Korea government (MSIT).

%
%
\bibliographystyle{splncs04}
\bibliography{main}
\end{document}